\documentclass{article}

 \usepackage[preprint]{neurips_2026}

\usepackage[utf8]{inputenc} 
\usepackage[T1]{fontenc}    
\usepackage{hyperref}       
\usepackage{url}            
\usepackage{booktabs}       
\usepackage{amsfonts}       
\usepackage{nicefrac}       
\usepackage{microtype}      
\usepackage{xcolor}         

\definecolor{mazecol}{HTML}{F2F1E7}
\definecolor{flowcol}{HTML}{37b5ac}
\definecolor{sokocol}{HTML}{81B29A}

\colorlet{mazetrans}{mazecol}
\colorlet{flowtrans}{flowcol!25}
\colorlet{sokotrans}{sokocol!50}

\usepackage{amsmath}

\usepackage{color}
\usepackage{colortbl}

\usepackage{booktabs}
\usepackage{multirow} 
\usepackage{soul}
\usepackage{xcolor} 
\usepackage{tabularx}
\usepackage{paralist}
\usepackage{diagbox}

\definecolor{lightgreen}{RGB}{199,249,204}
\definecolor{lightblue}{RGB}{189,224,254}

\definecolor{color1}{HTML}{0a9396}
\definecolor{color2}{HTML}{94d2bd}
\definecolor{color3}{HTML}{e9d8a6}
\definecolor{color4}{HTML}{ee9b00}
\definecolor{color5}{HTML}{ca6702}
\definecolor{color6}{HTML}{bb3e03}

\newcolumntype{x}{>{\columncolor{color1!30}}c}
\newcolumntype{o}{>{\columncolor{color2!30}}c}
\newcolumntype{y}{>{\columncolor{color3!30}}c}
\newcolumntype{g}{>{\columncolor{color4!30}}c}
\newcolumntype{q}{>{\columncolor{color5!30}}c}
\newcolumntype{z}{>{\columncolor{color6!30}}c}

\usepackage{pifont}
\usepackage{bbm}

\usepackage{subcaption}
\usepackage{graphicx}

\usepackage{wrapfig}
\definecolor{caribbeangreen}{rgb}{0.0, 0.8, 0.6}

\usepackage{enumitem}
\usepackage{tcolorbox}
\usepackage{lipsum} 

\newcommand{\stitle}[1]{\vspace{0em} \noindent{\bf #1.}}

\usepackage[capitalize]{cleveref}
\crefname{section}{Sec.}{Secs.}
\Crefname{section}{Section}{Sections}
\Crefname{table}{Table}{Tables}
\crefname{table}{Tab.}{Tabs.}

\title{Video Models Can Reason with Verifiable Rewards}

\author{
    Tinghui Zhu\textsuperscript{\rm $\spadesuit$} \quad
    Sheng Zhang\textsuperscript{\rm $\clubsuit$} \quad
    James Y. Huang\textsuperscript{\rm $\heartsuit$} \quad
    Selena Song\textsuperscript{\rm $\diamondsuit$} \\
    \textbf{Xiaofei Wen}\textsuperscript{\rm $\spadesuit$} \quad
    \textbf{Yuankai Li}\textsuperscript{\rm $\spadesuit$} \quad
    \textbf{Hoifung Poon}\textsuperscript{\rm $\clubsuit$} \quad
    \textbf{Muhao Chen}\textsuperscript{\rm $\spadesuit$}\\
\textsuperscript{\rm $\spadesuit$}{\small University of California, Davis} \quad
\textsuperscript{\rm $\clubsuit$}{\small Microsoft Research} \\
\textsuperscript{\rm $\heartsuit$}{\small University of Southern California} \quad
\textsuperscript{\rm $\diamondsuit$}{\small University of California, Santa Cruz}\\
\;{\small \texttt{\{thuzhu, muhchen\}@ucdavis.edu}} \quad
{\small\texttt{shezhan@microsoft.com}}\\
\vspace{-2mm}
\\
\small{\texttt{Project Page: }\url{https://darthzhu.github.io/VideoRLVR-page/}}
\vspace{-6mm}
}

\begin{document}

\maketitle

\begin{abstract}
Video diffusion models have made rapid progress in perceptual realism and temporal coherence, but they remain primarily optimized for plausible generation rather than verifiable reasoning. 
This limitation is especially pronounced in tasks where generated videos must satisfy explicit spatial, temporal, or logical constraints. 
Inspired by the role of reinforcement learning with verifiable rewards (RLVR) in reasoning-oriented language models, we introduce \textbf{VideoRLVR}, a practical recipe for optimizing video diffusion models with rule-based feedback. 
VideoRLVR formulates video reasoning as the generation of verifiable visual trajectories and consists of an SDE-GRPO optimization backbone, dense decomposed rewards, and an \textit{Early-Step Focus} strategy for efficient training. 
The Early-Step Focus strategy restricts policy optimization to the early denoising phase, reducing training latency by about 40\% while preserving performance. 
We evaluate VideoRLVR on Maze, FlowFree, and Sokoban, three procedurally generated domains with objective success criteria. 
Across these tasks, VideoRLVR consistently improves over supervised fine-tuning baselines, with dense decomposed rewards proving especially important in low-success-rate settings. 
Our RL-optimized model also outperforms the evaluated proprietary and open-source video generation models on these verifiable reasoning benchmarks and out-of-domain benchmarks. 
These results suggest that verifiable RL can move video models beyond perceptual imitation toward more reliable rule-consistent visual reasoning.
\end{abstract}
\section{Introduction}
\label{sec:introduction}

Recent progress in large language models (LLMs) has reshaped the role of generative models from content producers into increasingly capable reasoning systems~\citep{guo2025deepseek,singh2025openai,comanici2025gemini}.
A key intuition behind this shift is that the model can externalize the problem-solving process by generating intermediate states rather than only a final answer.
This raises a natural question for video generation: \emph{if language models can reason through sequences of tokens, can video models reason through sequences of frames?} 
Videos provide an appealing foundation for this idea, where each frame can represent an intermediate visual state in a goal-directed process. 
In domains such as navigation~\citep{dong2026language}, puzzle solving~\citep{hossieni2023puzzlefusion}, and embodied planning~\citep{mei2026video}, a generated video can therefore be viewed not merely as motion synthesis, but as 
a temporally ordered chain of visual states~\citep{wiedemer2025video} that encodes a visual reasoning trajectory.

\begin{figure}
    \centering
    \includegraphics[width=\linewidth]{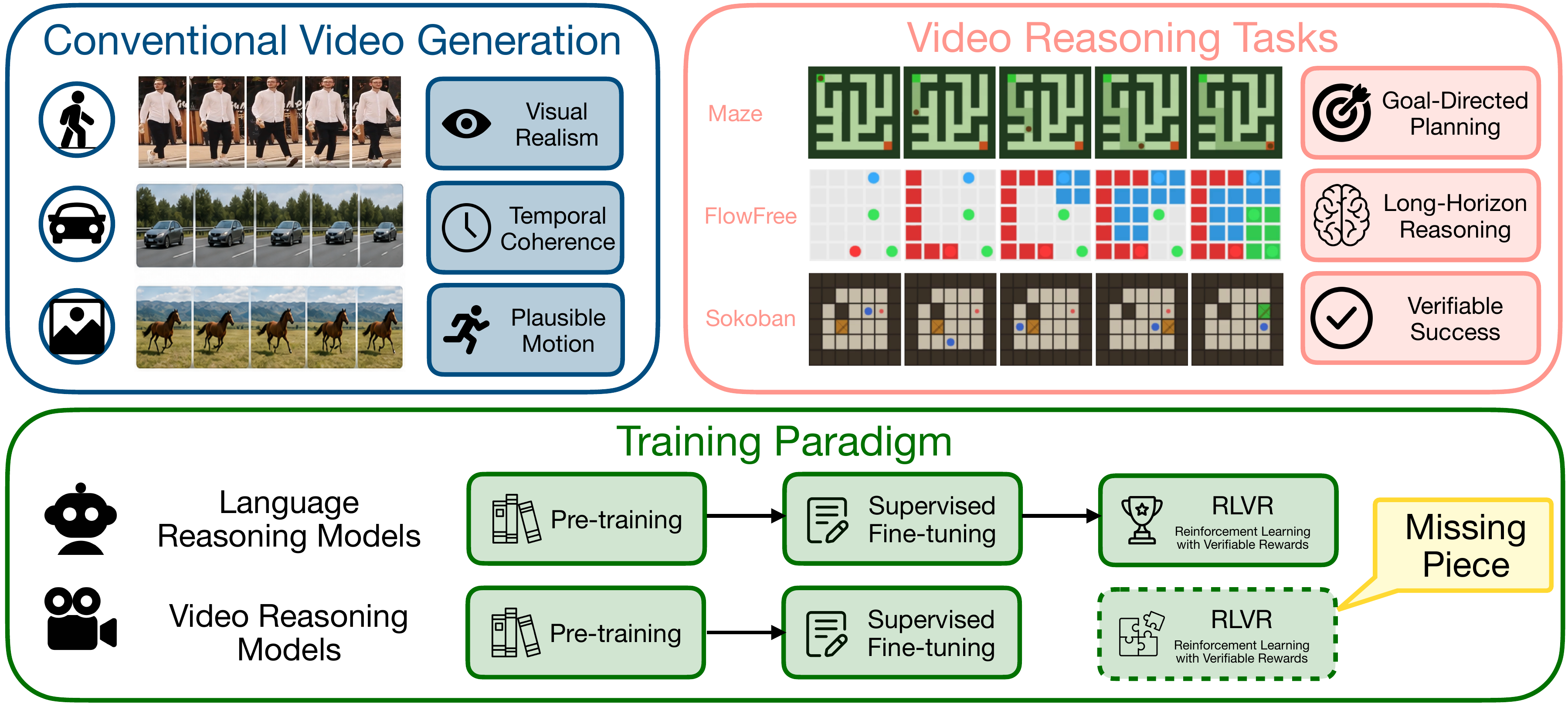}
    \vspace{-4mm}
    \caption{\textbf{Evolution towards verifiable video reasoning.} \textit{Top:} Comparison between perception-focused generation and reasoning-intensive tasks. \textit{Bottom:} We introduce \textbf{VideoRLVR}, the missing puzzle in the training paradigm for video reasoning models.}
    \label{fig:teaser}
    \vspace{-4mm}
\end{figure}

Despite this potential, current video diffusion models are still primarily optimized for perceptual quality, temporal coherence, and plausible motion~\citep{hong2022cogvideo,yang2023diffusion,wan2025wan}. 
While large-scale video models have begun to show signs of visual reasoning~\citep{wiedemer2025video,guo2025video,wang2026very}, these abilities remain difficult to elicit reliably and verify under standard training objectives. 
The core challenge is the mismatch between perceptual plausibility and objective correctness. 
Supervised fine-tuning (SFT) on ground-truth solution videos can teach the model the visual form of valid trajectories, yet it does not directly optimize the correctness of sampled outputs.
As a result, models may imitate solution-like patterns while failing to satisfy the underlying rules that make those solutions valid~\citep{geirhos2020shortcut,motamed2026generative}. 
This suggests an analogy to reasoning-oriented LLMs where pre-training provides broad generative competence, SFT teaches the format of reasoning traces, Reinforcement Learning with Verifiable Rewards (RLVR) is the essential third stage required to optimize objective correctness, as illustrated in \Cref{fig:teaser}.

In this work, we introduce \textbf{VideoRLVR}, a systematic recipe for applying reinforcement learning with verifiable rewards to video models. 
Our framework has three main components. 
First, we adopt an SDE-GRPO optimization backbone~\citep{liu2025flow} for optimizing flow-matching video models.
Second, we propose an \textit{Early-Step Focus} strategy for efficient video RL.
Instead of applying stochastic exploration and backpropagation across the entire denoising trajectory, this strategy concentrates optimization on the early denoising phase, where coarse structure and long-range planning are largely determined~\citep{wang2026demystifing}. 
Finally, we design dense decomposed rewards that break sparse task success into verifiable structural components, providing informative feedback even when full success is rare. 
To acquire dense reward signals, we construct verifiable video reasoning data by generating solution trajectories with rule-based planners and aligning each logical transition with the video frame sequence.

We evaluate our RLVR recipe on a multi-task suite designed for rule-based verification, including Maze, FlowFree, and Sokoban.
Our experiments show that VideoRLVR improves video reasoning beyond supervised imitation. 
Across all three domains, the RL-optimized model consistently achieves higher success rates than the SFT checkpoint used to initialize training, with gains of 6.1\%, 5.5\%, and 3.2\% on Maze, FlowFree, and Sokoban, respectively.
Compared with continued supervised training, VideoRLVR yields larger gains on harder tasks, suggesting that verifiable rewards provide an optimization signal 
beyond what can be captured by imitation alone.
We further evaluate VideoRLVR on the out-of-domain split of VBVR~\citep{wang2026very}, where VideoRLVR shows improved transfer beyond the training domains. 
Our ablations further show that dense decomposed rewards are crucial in low-success-rate domains, and that Early-Step Focus reduces training time by about 40\% 
while maintaining nearly the same performance.
Finally, VideoRLVR outperforms several proprietary and open-source video generation models on our verifiable reasoning benchmarks, indicating that targeted verifiable RL can substantially improve the logical correctness of generated visual trajectories.

In summary, our contributions are as follows:
\vspace{-2mm}
\begin{enumerate}[leftmargin=1.5em,itemsep=0pt]
    \item We introduce \textbf{VideoRLVR}, a reinforcement learning framework that optimizes video diffusion models with verifiable rewards, including dense decomposed reward functions to provide informative feedback for rule-verifiable visual trajectories.

    \item We introduce a scalable training pipeline that combines rule-based trajectory generation, SDE-GRPO optimization, and an Early-Step Focus strategy that reduces training time by about 40\% while preserving the performance.

    \item We show that VideoRLVR improves over supervised fine-tuning and competitive proprietary and open-source video generation models on Maze, FlowFree, and Sokoban, while also demonstrating improved out-of-domain transfer on VBVR.
\end{enumerate}
\vspace{-2mm}

\section{Related Work}
\label{sec:related_work}

\paragraph{Reinforcement learning for diffusion and flow-matching models.}
Reinforcement learning has increasingly been used to align diffusion and flow-based generative models with human preferences, perceptual objectives, and task-specific rewards~\citep{xue2026systematic}. 
Prior work formulates denoising as a sequential decision process and applies policy-gradient or preference-optimization methods to improve text-to-image and video generation~\citep{black2023training,fan2023dpok,wallace2024diffusion}. 
For flow-matching models, recent methods address the deterministic nature of ODE sampling by introducing stochastic transitions or alternative preference objectives, enabling likelihood-ratio or GRPO-style optimization~\citep{liu2025flow,xue2025dancegrpo,chen2024dgpo,mcallister2025flow}. 
Other extensions apply these ideas to video or embodied objectives~\citep{an2026vggrpo,liu2024diff}. 
However, existing work optimizes perceptual or preference-based criteria such as aesthetics, text rendering, image fidelity, geometric consistency, or motion quality~\citep{li2025growing,li2025branchgrpo}. 
In contrast, our work studies reinforcement learning for verifiable video reasoning, where rewards are computed from objective task rules and success depends on the logical correctness of the generated visual trajectory.

\paragraph{Reasoning in video generation models.}
Recent work has begun to investigate whether video generation models can serve as reasoning systems rather than only visual synthesizers. 
Large-scale video models have shown emerging abilities on visual puzzles and sequential prediction tasks, motivating the view that video generation can be interpreted as a chain of visual states or ``chain of frames''~\citep{wiedemer2025video,guo2025video,huang2025vchain}. 
Benchmark efforts~\citep{wang2026very,cai2025mmgr,yang2025reasoning,tong2025thinking} further evaluate video models on reasoning-oriented tasks that require temporal consistency, spatial planning, or rule satisfaction. 
Other studies analyze video models as world simulators or physical reasoners, highlighting both their potential and their limitations in capturing causal and physical structure~\citep{brooks2024video,kang2024far,mei2026video,motamed2026generative,zhang2025morpheus,song2025learning}. 
These works suggest that video models may contain useful visual reasoning priors, but also show that standard generation objectives do not reliably produce rule-correct trajectories~\citep{guo2025video,luo2025v}. 
Our work addresses this gap by directly optimizing video models with verifiable rewards, using rule-based success criteria rather than relying solely on supervised imitation or zero-shot generation.

\paragraph{Verifiable reinforcement learning and reasoning models.}
Reinforcement learning with verifiable rewards has played an important role in recent progress on reasoning-oriented language models~\citep{guo2025deepseek,singh2025openai,comanici2025gemini}. 
In these settings, the model is rewarded according to objective correctness signals, such as mathematical equivalence, executable code tests, or rule-based verification, instead of only human preference judgments~\citep{li2025system,zeng2025simplerl,hu2025open,huang2026learning}. 
This paradigm is attractive because it provides scalable supervision when outcomes can be automatically checked, which facilitates the development of emerging behaviors like searching and backtracking~\citep{zhu2024deductive,wu2025arm}.
Our work extends this training from language outputs to video trajectories. 
Whereas text reasoning is often verified by final-answer correctness, video reasoning requires trajectory-level verification over visual, temporal, and process constraints. 
We study how verifiable RL can optimize video diffusion models under these criteria.

\section{Problem Formulation}
\label{sec:problem_formulation}

\stitle{RLVR for Video Reasoning}
Following \cite{wiedemer2025video}, we formulate video reasoning as a conditional generation task where a model generates a temporal sequence of visual states whose transitions and terminal state can be checked against task-specific rules.
Given an initial image $I_0$ and a textual instruction $T$, let $c=(I_0,T)$ denote the conditioning input.
The model generates a video $\mathbf{V}=\{I_0,I_1,\ldots,I_{F-1}\}$, where $F$ is the number of frames.
Unlike standard video synthesis, which primarily evaluates perceptual quality and temporal coherence, video reasoning requires the generated sequence to satisfy task-specific correctness criteria.
This formulation allows us to treat video generation as a search for a valid visual trajectory conditioned on the initial state and instruction.

\stitle{Video Generation as a Markov Decision Process}
To apply reinforcement learning to flow-matching video generation, we formulate the reverse denoising process as a Markov Decision Process (MDP) over latent variables.
This MDP is defined over denoising steps rather than reasoning steps, where the reward is computed after the final video is decoded.
At denoising step $k \in \{1,\ldots,K\}$, the state is the noisy video latent $x_{t_k}$ at noise level $t_k$, and the action is the model velocity prediction $\hat{v}_\theta(x_{t_k},t_k,c)$, which determines the mean update of the next latent.
Under the Ordinary Differential Equation (ODE) solver, the transition is given by $x_{t_{k+1}} = x_{t_k} + (t_{k+1}-t_k)a_k .$
After the final denoising step, the decoded video $\mathbf{V}$ receives a verifier-derived reward $R(\mathbf{V},c)$.
A fundamental challenge in this formulation is that standard flow matching employs a deterministic ODE solver, making it a deterministic function of the initial noise $x_1$.
Under this deterministic solver, the next latent is a deterministic function of $(x_{t_k}, c)$, yielding no tractable stochastic transition density $\pi_\theta(x_{t_{k+1}}\mid x_{t_k}, c)$ for likelihood-ratio policy gradients.
In~\Cref{sec:methods}, we address this by adopting an SDE-based formulation that introduces stochastic transitions compatible with flow-matching generation.

\stitle{Tasks}
To evaluate VideoRLVR across different reasoning domains, we instantiate our framework on three rule-verifiable visual reasoning domains: Maze, FlowFree, and Sokoban.
We choose these tasks because they satisfy three properties: 
1) solution correctness can be checked by rule-based verifiers, 
2) large-scale training and test instances can be generated, and 
3) the tasks span different levels of reasoning complexity.
Maze primarily tests spatial connectivity under explicit obstacle constraints, FlowFree requires globally consistent non-overlapping path connectivity and implicit constraints, and Sokoban introduces object interaction, irreversible transitions, and longer-horizon reasoning.

\section{RLVR Recipe for Video Reasoning Models}
\label{sec:methods}

We present \textbf{VideoRLVR}, a systematic recipe for optimizing video models with verifiable rewards. 
The recipe consists of three components: 
1) an SDE-GRPO optimization backbone, 
2) an Early-Step Focus optimization strategy, and
3) dense decomposed rewards design and acquisition.

\subsection{SDE-GRPO for Video Reasoning}
\label{sec:sde_grpo}
GRPO~\citep{shao2024deepseekmath} estimates relative advantages from groups of sampled outputs without training a separate critic, making it well suited for verifiable reward settings.
However, standard flow-matching models generate samples with a deterministic ODE sampler, which does not provide a tractable stochastic transition density over denoising steps.
Following Flow-GRPO~\citep{liu2025flow}, we convert the deterministic denoising dynamics into stochastic transitions with Gaussian log-probabilities.

\stitle{Stochastic denoising transitions}
For a discretized denoising schedule $\{t_k\}_{k=1}^{K}$, the SDE formulation defines a Gaussian transition:
\begin{equation}
\label{eq:transition}
\pi_\theta(x_{t_{k+1}} \mid x_{t_k}, c)
=
\mathcal{N}\!\left(
x_{t_{k+1}};
\mu_\theta(x_{t_k},t_k,c),
\sigma_k^2\mathbf{I}
\right),
\end{equation}
where $\mu_\theta(x_{t_k},t_k,c)$ is the mean update induced by the model and $\sigma_k^2$ is the SDE transition variance.
This stochastic transition enables closed-form log-probabilities and likelihood-ratio policy gradients.

\stitle{GRPO objective}
Given a group of $G$ sampled videos for each condition, we compute verifier-derived rewards and normalize them within the group to obtain advantages $A_i$.
For each sample $i$ and denoising step $k$, we compute the dimension-normalized log-ratio:
\begin{equation}
\label{eq:log_ratio}
\begin{aligned}
\log \rho_{i,k}
&=
\log
\frac{
    \pi_\theta(x^{(i)}_{t_{k+1}} \mid x^{(i)}_{t_k}, c_i)
}{
    \pi_{\text{old}}(x^{(i)}_{t_{k+1}} \mid x^{(i)}_{t_k}, c_i)
}
=
-\frac{1}{2\sigma_k^2}
\cdot
\frac{1}{D}
\sum_{d=1}^{D}
\left[
\left(x^{(i)}_{t_{k+1}}-\mu^{(i,k)}_\theta\right)_d^2
-
\left(x^{(i)}_{t_{k+1}}-\mu^{(i,k)}_{\text{old}}\right)_d^2
\right],
\end{aligned}
\end{equation}
where $\mu^{(i,k)}_\theta=\mu_\theta(x^{(i)}_{t_k},t_k,c_i)$, 
$\mu^{(i,k)}_{\mathrm{old}}=\mu_{\mathrm{old}}(x^{(i)}_{t_k},t_k,c_i)$, and $D$ is the number of latent elements.
The policy loss uses PPO-style clipping:
\begin{equation}
\label{eq:ppo_loss}
\mathcal{L}_{\mathrm{policy}}
=
-\mathbb{E}_{i,k}
\left[
\min\left(
\rho_{i,k}A_i,\,
\mathrm{clip}(\rho_{i,k},1-\varepsilon,1+\varepsilon)A_i
\right)
\right].
\end{equation}
We additionally regularize the policy against the reference model with a closed-form KL penalty:
\begin{equation}
\label{eq:kl}
\mathcal{L}_{\mathrm{KL}}
=
\mathbb{E}_k
\left[
\frac{1}{D}
\frac{\|\mu_\theta-\mu_{\mathrm{ref}}\|_2^2}{2\sigma_k^2}
\right].
\end{equation}
The final objective is $\mathcal{L}_{\text{VideoRLVR}}=\mathcal{L}_{\text{policy}}+\beta \mathcal{L}_{\text{KL}},$where $\beta$ controls the strength of regularization.

\subsection{Early-Step Focus for Efficient Video RL}
\label{sec:early_step_focus}

Video RL is substantially more expensive than text RL because each rollout requires generating and backpropagating through high-dimensional spatio-temporal latents.
A full SDE-GRPO update over all $K$ denoising steps therefore incurs large memory and time costs.
However, not all denoising steps contribute equally to the reasoning objective.
Early high-noise steps are primarily responsible for coarse layout, object placement, and long-range structure, whereas later low-noise steps mainly refine local appearance and consolidate the generation into a specific visual trajectory~\citep{wang2026demystifing}.

Motivated by this observation, we introduce \textit{Early-Step Focus}.
During RL optimization, we sample the full denoising trajectory for generation and reward evaluation, but restrict stochastic perturbation, log-probability computation, and gradient backpropagation to the first $L<K$ denoising steps.
This creates an efficient exploration-exploitation trade-off: early denoising steps receive stochastic perturbations and policy-gradient updates for high-level reasoning, while later steps preserve the generative prior and refine visual details.
The policy loss becomes:
\begin{equation}
\label{eq:esf_loss}
\mathcal{L}_{\text{ESF}}
=
-\mathbb{E}_{i,k\leq L}
\left[
\min\left(
\rho_{i,k} A_i,\,
\mathrm{clip}(\rho_{i,k},1-\varepsilon,1+\varepsilon)A_i
\right)
\right]
+
\beta \mathcal{L}_{\text{KL}}^{k\leq L}.
\end{equation}
In our experiments, we use $K=20$ denoising steps and $L=10$ early steps.
This reduces training latency by about 40\% while preserving reasoning performance, suggesting that the early denoising phase carries most of the reward-relevant structural signal.

\subsection{Verifiable Reward Design and Acquisition}
\label{sec:reward_design}

A key requirement for VideoRLVR is that generated videos can be automatically parsed and evaluated.
Existing video reasoning datasets~\citep{yang2025reasoning,wang2026very} often lack the scale, task diversity, or fine-grained difficulty variation required to study RLVR for video reasoning.
We synthesize task instances with rule-based planners that sample an initial configuration, solve it with a valid action sequence, and render the resulting state trajectory into a video.
Alongside each trajectory, we retain environment metadata, such as grid layouts, endpoint locations, object states, and goal conditions, which is used for automatic verification and reward computation.
Each discrete environment action is mapped to a unique frame transition $I_f \to I_{f+1}$, making the generated video directly interpretable as a reasoning trajectory.
Task-specific generation details are provided in~\Cref{sec:appd_dataset}.

Given the metadata from the data curation process, we now can convert task rules into dense reward signals.
Instead of using only a binary success reward signal, we decompose each task into structural components that measure partial progress toward a valid solution.
This is especially important in low-success-rate domains, where most sampled videos receive zero reward and therefore provide little variation within a GRPO group.

\stitle{Task-aware Reward Function}
We use a task-aware reward function for joint training across heterogeneous domains.    
For each conditioning input $c$, the dispatcher identifies the task $\mathcal{T}(c)\in\{\text{Maze},\text{FlowFree},\text{Sokoban}\}$ and evaluates the generated video with the corresponding reward:
\vspace{-1mm}
\begin{equation}
R(\mathbf{V}, c) = R_{\mathcal{T}(c)}(\mathbf{V}, c).
\end{equation}
This allows mixed-task RL batches while preserving task-specific verification criteria.

\stitle{Dense Reward Formulations}
For each task, we decompose the global objective into measurable rule-based components:
\vspace{-2mm}
\begin{itemize}[leftmargin=1.5em,itemsep=0pt]
    \item \textbf{Maze.}
    We define the reward as: $R_{\text{maze}} = R_{\text{conn}} \cdot R_{\text{wall}},$
    where $R_{\text{conn}}$ measures start-to-goal path connectivity and $R_{\text{wall}}$ penalizes wall violations.
    Compared with an additive formulation, the multiplicative form produces sharper reward separation within a GRPO group by assigning high scores only to trajectories that satisfy both connectivity and wall consistency, yielding more informative relative advantages.

    \item \textbf{FlowFree.}
    We combine four structural metrics: $R_{\text{ff}} =
    \lambda_{\text{valid}} R_{\text{valid}}
    + \lambda_{\text{pres}} R_{\text{pres}}
    + \lambda_{\text{conn}} R_{\text{conn}}
    + \lambda_{\text{fill}} R_{\text{fill}},$
    where $R_{\text{valid}}$ measures endpoint-to-endpoint path validity, $R_{\text{pres}}$ measures preservation of the given endpoints, $R_{\text{conn}}$ measures 4-connected color regions, and $R_{\text{fill}}$ measures grid coverage by valid path colors.
    The weights $\lambda_{\text{valid}}, \lambda_{\text{pres}}, \lambda_{\text{conn}}, \lambda_{\text{fill}}$ balance the relative importance of these components.
    In our experiments, we set them to be $0.15$, $0.35$, $0.30$, and $0.20$, respectively.

    \item \textbf{Sokoban.}
    We use a combination of final-state and process-validity rewards: $R_{\text{sok}} =
    \lambda_{\text{state}} R_{\text{state}}
    +
    \lambda_{\text{proc}} R_{\text{proc}},$
    where $R_{\text{state}}$ measures box placement on target cells and $R_{\text{proc}}$ measures the fraction of valid transitions under Sokoban movement rules.
    The weights $\lambda_{\text{state}}$ and $\lambda_{\text{proc}}$ balance final-state correctness and process validity.
    We use $\lambda_{\text{state}}=\lambda_{\text{proc}}=0.5$ in all experiments.
\end{itemize}

\section{Experiments}
\label{sec:experiments}

In this section, we evaluate \textbf{VideoRLVR} from two perspectives.
First, we compare against supervised fine-tuning and competitive video generation baselines on three rule-verifiable reasoning domains: Maze, FlowFree, and Sokoban.
Then, we test transfer beyond the training domains using the out-of-domain split of VBVR~\citep{wang2026very}.
Together, these experiments assess whether verifiable RL improves both in-domain rule-based correctness and out-of-domain visual reasoning behavior.

\subsection{Experimental Setup}

\stitle{Dataset}
We train and evaluate on a multi-task suite of three procedurally generated reasoning domains: Maze, FlowFree, and Sokoban.
To prevent the model from overfitting to specific visual features, we apply varied color themes across the dataset, encouraging the model to rely on structural invariants.
Each sample consists of an input image, a task instruction, and an 81-frame ground-truth video at 480${\times}$832 resolution.
The total training dataset consists of 30,000 samples (10,000 per task).
For the test set, we maintain a held-out set of 3,000 samples (1,000 per task) generated with disjoint random seeds.
Dataset construction details are provided in~\Cref{sec:appd_dataset_setup}.

\stitle{Base Model and SFT Baseline}
We use Wan2.2-TI2V-5B~\cite{wan2025wan}, a state-of-the-art video generation model, as our base model.
It generates $F=81$ frames at $480 \times 832$ resolution.
We first establish an SFT baseline by training the model on ground-truth solution videos using the standard flow matching objective.
This SFT checkpoint provides the necessary perceptual and structural prior for the model, serving as both the initial policy and the reference policy $\pi_{\text{ref}}$ for RL optimization.

\stitle{Baselines}
To evaluate the effectiveness of our method, we compare our model with competitive proprietary and open-source video generation baselines.
For proprietary models, we use Sora 2~\citep{sora2_system_card_openai_2025}, Kling V3~\citep{kling_video_3_0_2026}, and Veo 3.1~\citep{wiedemer2025video}.
For open-source models, we compare with Wan2.2-TI2V-5B~\citep{wan2025wan}, CogVideoX1.5-5B-I2V~\citep{hong2022cogvideo}, and HunyuanVideo-I2V~\citep{wu2025hunyuanvideo}.
We also compare with specialized SFT-based video reasoning models, including Wan-R1~\citep{yang2025reasoning} and VBVR-Wan2.2~\citep{wang2026very}.
Wan-R1 adopts the same base model as ours and is trained on the Maze and Sokoban domains with LoRA~\citep{hu2022lora}.
VBVR-Wan2.2 utilizes Wan2.2-I2V-A14B~\citep{wan2025wan} and is trained on the VBVR dataset with LoRA.

\stitle{Training Configuration}
We train the SFT baseline for 5 epochs with a learning rate of $1{\times}10^{-5}$.
For VideoRLVR, we initialize from the SFT checkpoint and train on the same training set for $1$ epoch using SDE-GRPO as the optimization backbone.
We use group size $G=16$, $T=20$ denoising steps, learning rate $5{\times}10^{-6}$, and KL coefficient $\beta=0.04$.
Following the Early-Step Focus strategy in~\Cref{sec:early_step_focus}, backpropagation and SDE injection are restricted to the first $L=10$ steps of the denoising trajectory.
All training experiments are conducted on $8$ NVIDIA B200 GPUs.

\stitle{Evaluation}
\label{sec:gt_eval}
We evaluate the results using two complementary metric families: 
1) trajectory alignment metrics, including Precision (Prec), Recall (Rec), and F1, which measure pixel-, cell-, or action-level alignment with the reference solution, and 
2) symbolic success rate, which verifies whether the video satisfies the underlying task rules.
Evaluation details are listed in \Cref{sec:appd_eval}

\begin{table}[t]
\small
\vspace{-5mm}
\caption{Comparison of our method with other baselines. We report Precision (Prec), Recall (Rec), F1, and Success Rate (SR). \textbf{Bold} indicates the best and \underline{underlined} indicates second best.}
\begin{tabular}{l >{\columncolor{mazetrans}}c >{\columncolor{mazetrans}}c >{\columncolor{mazetrans}}c >{\columncolor{mazetrans}}c >{\columncolor{flowtrans}}c >{\columncolor{flowtrans}}c >{\columncolor{flowtrans}}c >{\columncolor{flowtrans}}c >{\columncolor{sokotrans}}c >{\columncolor{sokotrans}}c >{\columncolor{sokotrans}}c >{\columncolor{sokotrans}}c}
\toprule
\multirow{2}{*}{Model}              & \multicolumn{4}{c}{\cellcolor{mazetrans}Maze}                                                                             & \multicolumn{4}{c}{\cellcolor{flowtrans}FlowFree}                                                                         & \multicolumn{4}{c}{\cellcolor{sokotrans}Sokoban}                                                                        \\ \cmidrule(lr){2-5}\cmidrule(lr){6-9}\cmidrule(lr){10-13}
                                    & \multicolumn{1}{c}{Prec} & \multicolumn{1}{c}{Rec} & \multicolumn{1}{c}{F1} & \multicolumn{1}{c}{SR} & \multicolumn{1}{c}{Prec} & \multicolumn{1}{c}{Rec} & \multicolumn{1}{c}{F1} & \multicolumn{1}{c}{SR} & \multicolumn{1}{c}{Prec} & \multicolumn{1}{c}{Rec} & \multicolumn{1}{c}{F1} & \multicolumn{1}{c}{SR} \\ \hline
\multicolumn{13}{c}{{\textit{Proprietary Models}}} 
\\
\hline
\multicolumn{1}{l|}{Sora 2}         & 15.8                     & 17.2                    & 16.5                   & 3.1                    & 10.8                     & 5.1                     & 5.8                    & 0.0                    & 8.5                    & 4.8                     & 5.4                    & 0.0                    \\
\multicolumn{1}{l|}{Kling V3}         & 24.8                     & 15.7                    & 19.2                   & 23.5                    & 18.8                     & 2.7                     & 4.7                    & 0.0                    & 5.7                    & 2.7                     & 3.7                    & 0.0                    \\
\multicolumn{1}{l|}{Veo 3.1}        & 22.8                     & 18.1                    & 20.2                   & 26.0                   & 23.9                     & 4.7                     & 7.5                    & \underline{4.0}                    & 22.2                   & 6.0                     & 9.4                    & 0.0                    \\ \hline
\multicolumn{13}{c}{{\textit{Open-Source Models}}}       \\                                                                                                            
\hline
\multicolumn{1}{l|}{CogVideoX1.5} & 13.3                     & 10.8                    & 11.9                   & 0.0                    & 18.7                     & 2.2                     & 3.9                    & 0.0                    & 3.2                    & 0.3                     & 0.5                    & 0.0                    \\
\multicolumn{1}{l|}{HunyuanVideo} & 17.3                     & 11.4                    & 13.8                   & 2.2                    & 12.5                     & 2.9                     & 4.8                    & 0.0                    & 8.2                    & 2.7                     & 3.2                    & 0.0                    \\
\multicolumn{1}{l|}{Wan2.2-TI2V-5B} & 18.3                     & 12.2                    & 14.6                   & 0.0                    & 17.4                     & 2.0                     & 3.4                    & 0.0                    & 4.1                    & 0.7                     & 1.0                    & 0.0                    \\
\hline
\multicolumn{13}{c}{{\textit{SFT Models}}} \\
\hline
\multicolumn{1}{l|}{Wan-R1} & 20.9                     & 65.6                    & 31.7                   & 31.9                    & 20.9                     & 3.6                     & 6.1                    & 0.0                    & 7.7                    & 2.1                     & 3.3                    & 0.0                    \\
\multicolumn{1}{l|}{VBVR-Wan2.2} & 62.7                     & 77.8                    & 69.4                   & 60.8                    & 17.9                     & 5.6                     & 8.5                    & 1.7                    & 16.2                    & 1.7                     & 3.1                    & 0.0                    \\
\multicolumn{1}{l|}{SFT Epoch 5}    & 80.2                     & 83.0                    & 81.6                   & 66.1                   & 42.8                     & 42.2                    & 42.4                   & 2.4                    & 33.6                   & 11.9                    & 17.6                   & \underline{2.9}                    \\
\multicolumn{1}{l|}{SFT Epoch 10}   & 80.4                     & 85.1                    & 82.7                   & \underline{69.0}                   & 43.1                     & 42.5                    & 42.8                   & 2.5                    & 32.8                   & 11.6                    & 17.1                   & 2.7                    \\
\hline
\multicolumn{13}{c}{{\textit{RL Model}}} \\
\hline
\multicolumn{1}{l|}{\textbf{VideoRLVR}}  & 82.1                     & 86.9                    & 84.4                   & \textbf{72.2}                   & 44.3                     & 43.8                    & 44.0                   & \textbf{7.9}                    & 34.0                   & 12.5                    & 29.4                   & \textbf{6.1}                    \\ \hline
\end{tabular}
\vspace{-5mm}
\label{tab:main}
\end{table} 

\subsection{Main Results}
\label{sec:main_results}
\Cref{tab:main} compares VideoRLVR with supervised baselines and competitive video generation models on our verifiable reasoning benchmarks.

\textbf{RLVR consistently outperforms supervised baselines.}
VideoRLVR yields consistent improvements across all three reasoning domains.
Compared with the SFT Epoch 5 checkpoint used to initialize RL training, VideoRLVR improves success rate by 6.1\% on Maze, 5.5\% on FlowFree, and 3.2\% on Sokoban.
Notably, VideoRLVR also significantly surpasses the performance of recent state-of-the-art closed-source models on visual reasoning tasks, validating the efficacy of verifiable reinforcement learning in domains where generic video pre-training remains insufficient for complex logical tasks.

\textbf{Superior scaling on high-complexity tasks.}
To isolate the specific advantages of RLVR over extended supervised learning, we evaluate a stronger SFT baseline (SFT Epoch 10), representing the result of conducting further supervised training on the same Epoch 5 checkpoint.
\begin{wrapfigure}[15]{r}{0.5\columnwidth}
    \centering
    \vspace{-2mm}
    \includegraphics[width=.5\columnwidth]{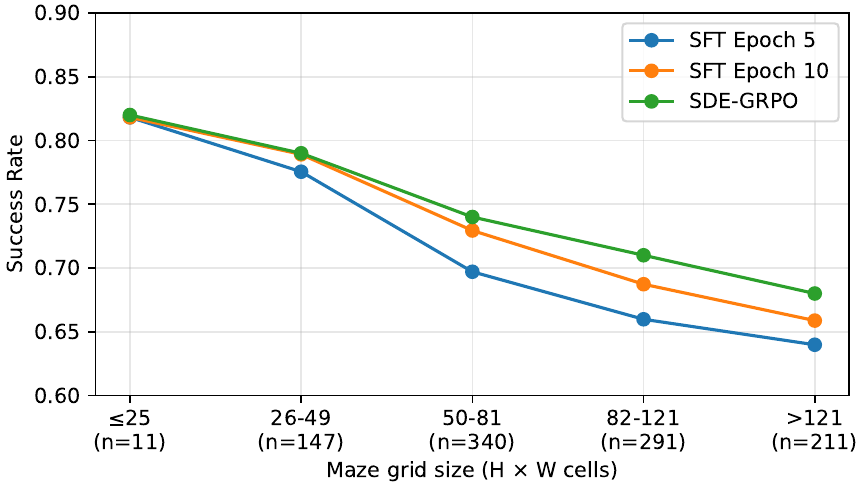}
    \vspace{-5mm}
    \caption{Success rate of different grid size for maze. $n$ represents the number of samples falling into this range}
    \label{fig:maze_sr_by_grid_size}
\end{wrapfigure}
As shown in \Cref{tab:main} and \Cref{fig:maze_sr_by_grid_size}, VideoRLVR is more robust than continued SFT as task difficulty increases.
Within the Maze domain, RLVR establishes a 3.2\% margin over the SFT Epoch 10 checkpoint and shows less degradation when the scale of the maze increases.
On FlowFree, VideoRLVR improves over SFT Epoch 10 by 5.4\%, while continued SFT provides little improvement over the Epoch 5 checkpoint.
On Sokoban, continued SFT slightly degrades performance, whereas VideoRLVR improves over SFT Epoch 10 by 3.4\%.
These trends suggest that verifiable rewards provide an optimization signal that is not captured by additional imitation training alone.



\begin{wraptable}[9]{r}{0.48\textwidth}
\small
\vspace{-4mm}
\caption{Comparison with LLMs on maze tasks.}
\vspace{-2mm}
\begin{tabular}{lcccc}
\toprule
\multirow{2}{*}{Model}                & \multicolumn{4}{c}{Maze}                                                                             \\ \cline{2-5} 
                                      & \multicolumn{1}{c}{Prec} & \multicolumn{1}{c}{Rec} & \multicolumn{1}{c}{F1} & \multicolumn{1}{c}{SR} \\ \hline
\multicolumn{1}{l|}{GPT 4o}           & 11.7                     & 13.0                    & 12.3                   & 0.0                    \\
\multicolumn{1}{l|}{GPT 5.5 Pro}      & 76.0                     & 70.1                    & 72.9                   & 66.0                   \\
\multicolumn{1}{l|}{Gemini 2.5 Flash} & 11.2                     & 10.5                    & 10.9                   & 0.0                    \\
\multicolumn{1}{l|}{Gemini 3.1 Pro}   & 26.8                     & 27.0                    & 26.9                   & 23.0                   \\
\multicolumn{1}{l|}{\textbf{VideoRLVR}}             & \textbf{82.1}                     & \textbf{86.9}                    & \textbf{84.4}                   & \textbf{72.2}                  \\
\bottomrule
\end{tabular}
\label{tab:llm_comparison}
\end{wraptable}
\subsection{Comparison with LLMs}
To determine if our reasoning domains can be solved by language reasoning alone, we benchmark frontier LLMs on the Maze task.
\Cref{tab:llm_comparison} presents the results of state-of-the-art models, including GPT-5.5 Pro~\citep{openai2026gpt55} and Gemini 3.1 Pro~\citep{googledeepmind2026gemini31pro}, compared against our RLVR-optimized video model.
Despite their sophisticated reasoning capabilities in textual domains, LLMs exhibit a sharp performance decay in maze tasks.
This divergence highlights a representation bottleneck: while LLMs must reason over a tokenized rendering of the maze, our video model operates directly on a visual latent space that inherently preserves the visual topological relationships necessary for complex visual reasoning.
These results suggest that, for visual reasoning, directly generating and optimizing visual trajectories can be more effective than solving the task through language-token representations alone.

\subsection{OOD Results}
\label{sec:ood_results}
\begin{wraptable}[10]{r}{0.7\textwidth}
\vspace{-4mm}
\caption{\textbf{OOD evaluation on VBVR.} We report average performance and category-wise scores on the VBVR-OOD split.}
\label{tab:vbvr}
\small
\vspace{-2mm}
\begin{tabular}{lcccccc}
\toprule
\textbf{Model}                    & \textbf{Avg.} & \textbf{Abst.} & \textbf{Know.} & \textbf{Perc.} & \textbf{Spat.} & \textbf{Trans.} \\
\hline
\multicolumn{7}{l}{\underline{\textit{5B Models}}}                                                                                    \\
CogVideoX1.5 & 26.2          & 28.1           & 23.5           & 25.0           & 25.4           & 28.2            \\
\textbf{VideoRLVR}  & \underline{60.2}    & \underline{65.5}     & \textbf{62.0}  & \textbf{59.7}  & \underline{58.8}     & \underline{58.2}      \\
\hline
\multicolumn{7}{l}{\underline{\textit{14B Models}}}                                                                                   \\
Wan2.2-I2V-A14B     & 32.9          & 40.5           & 30.8           & 34.3           & 23.6           & 30.7            \\
VBVR-Wan2.2         & \textbf{61.0} & \textbf{76.8}  & \underline{57.2}     & \underline{54.7}     & \textbf{61.8}  & \textbf{61.5} \\
\bottomrule
\end{tabular}
\end{wraptable}
To evaluate if VideoRLVR transfers beyond the training domains, we test our model on the out-of-domain split of VBVR~\citep{wang2026very}. 
This benchmark covers multiple reasoning categories and therefore provides a broader test of whether VideoRLVR improves general video reasoning behavior beyond Maze, FlowFree, and Sokoban.
As shown in~\Cref{tab:vbvr}, VideoRLVR substantially improves over the 5B baseline, increasing the average score from 26.2 to 60.2 with gains across all VBVR-OOD categories.
VideoRLVR also performs competitively with the larger 14B VBVR-Wan2.2 model, achieving a similar average score despite using a smaller 5B backbone and much less training data. 
These results suggest that VideoRLVR learns transferable visual reasoning ability that generalizes beyond the generated training tasks.

\section{Analysis}
In this section, we analyze the main components of VideoRLVR, including GRPO group size, Early-Step Focus, KL regularization, dense reward design, and qualitative generation behavior.

\subsection{Ablation Study}

\stitle{Impact of Group Size}
In GRPO, the group size $G$ affects the stability of group-relative advantage estimation.
We investigate the impact of this hyperparameter within the Maze reasoning
\begin{wrapfigure}[12]{r}{0.5\columnwidth}
    \centering
    \vspace{-4mm}
    \includegraphics[width=.5\columnwidth]{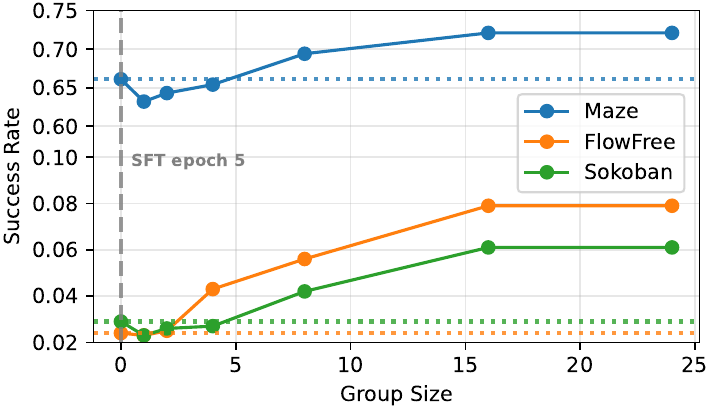}
    \vspace{-6mm}
    \caption{Scaling results with group size.}
    \label{fig:group_size}
\end{wrapfigure}
domain, as shown in \Cref{fig:group_size}.
Our results indicate that performance scales positively with group size, primarily due to the stabilization of the reward distribution's statistics.
While the expected sample standard deviation of rewards is a property of the policy's diversity, a small group size (e.g., $G \leq 4$) provides a noisy and often biased estimate of this value.
This leads to significant fluctuations in the advantage calculation, as the group mean fails to accurately represent the current policy's performance level.
Increasing the group size to $G=16$ provides a more stable comparison set for estimating relative advantages, which improves training stability in our experiments.
However, we observe diminishing returns beyond this point.
Furthermore, because video generation remains a significant computational bottleneck, scaling $G$ entails a linear increase in rollout time and VRAM overhead.
We therefore use $G=16$ as a practical trade-off between advantage-estimation stability and computational cost.

\stitle{Early-Step Focus}
To validate the efficacy of our Early-Step Focus strategy, we conduct a controlled experiment within the Maze domain.
We fix the total inference budget at $T=20$ denoising steps
\begin{wraptable}[6]{r}{0.5\textwidth}
\vspace{-3.5mm}
\caption{Comparisons of computing over full denoising steps and Early-Step Focus on Maze.}
\label{tab:gradient_step}
\vspace{-2mm}
\small
\begin{tabular}{l|ccc}
\toprule
Gradient Steps      & F1   & SR   & Time / step \\
\hline
20 (Full)           & 84.6 & 72.3 & 156 s       \\
10 (Early 10 steps) & 84.4 & 72.2 & 93.5 s \\
\bottomrule
\end{tabular}
\end{wraptable}
and compare the reasoning performance when the gradient and noise injection is calculated over the full trajectory ($L=20$) versus the first $L=10$ steps.
As illustrated in~\Cref{tab:gradient_step}, the success rates and F1 scores remain nearly unchanged, while training time is substantially reduced.
This suggests that the early denoising steps carry much of the reward-relevant structural signal for visual reasoning.
Because the later denoising steps primarily govern local textural refinement, they contribute less to the verifier-derived reasoning objective in this setting.
Restricting backpropagation and noise injection to these early steps thus serves as a computationally efficient optimization path, significantly reducing the training time without degrading the performance.

\stitle{KL Constraint}
The KL-divergence constraint is essential for maintaining the model's generative prior.
We show a qualitative example in~\Cref{sec:appd_kl}.
As shown in~\Cref{fig:KL_ablation}, removing the KL penalty ($\beta=0$) at an early stage of GRPO optimization can lead to reward-hacking behavior.
Without this regularization, the model may produce visually implausible patterns that satisfy parts of the verifier while degrading generation quality.
Implementing a constant penalty of $\beta=0.04$ successfully anchors the optimization to the original quality, ensuring that improvements in logical success are achieved without sacrificing the model's inherent visual plausibility.

\subsection{Reward Design}
\begin{wraptable}[5]{r}{0.55\textwidth}
\small
\vspace{-4mm}
\caption{Training with a sparse binary success reward.}
\label{tab:pure_success}
\vspace{-2mm}
\begin{tabular}{lllllll}
\toprule
Steps & 0    & 200  & 400  & 600  & 800  & 1000 \\
\hline
Maze     & 66.1 & 67.2 & 68.9 & 70.1 & 71.5 & 72.9 \\
FlowFree & 2.4  & 2.3  & 2.5  & 2.4  & 2.6  & 2.5 \\
\bottomrule
\end{tabular}
\end{wraptable}
To evaluate the necessity of our dense decomposed reward design, we investigate the efficacy of a sparse reward ($R \in \{0, 1\}$) based exclusively on success rate.
This ablation aims to determine if binary feedback is sufficient across varying levels of task complexity.

Our results, shown in \Cref{tab:pure_success}, reveal that sparse success rewards behave differently across domains.
In domains like maze, where the baseline model already achieves a decent success rate, the sparse reward signal is sufficient to provide an informative gradient.
The model is able to encounter success frequently enough during group rollouts to differentiate between advantageous and disadvantageous trajectories.
Conversely, on high-complexity tasks like FlowFree and Sokoban, where the initial success rate is near-zero, the sparse reward provides little useful signal.
In these environments, the model suffers from extreme gradient sparsity.
Since success is rarely encountered within the group rollout $G$, the advantage estimates remain uninformative.
This underscores a critical cold-start problem in video RL: while binary success is the ultimate goal, it is an insufficient signal for exploring high-dimensional latent space from a low-performance starting point.
Dense decomposed rewards address this issue by providing partial credit for intermediate structural properties, giving the policy useful feedback before full success becomes frequent.

\begin{figure}[t]
    \centering
    \vspace{-4mm}
    \includegraphics[width=\linewidth]{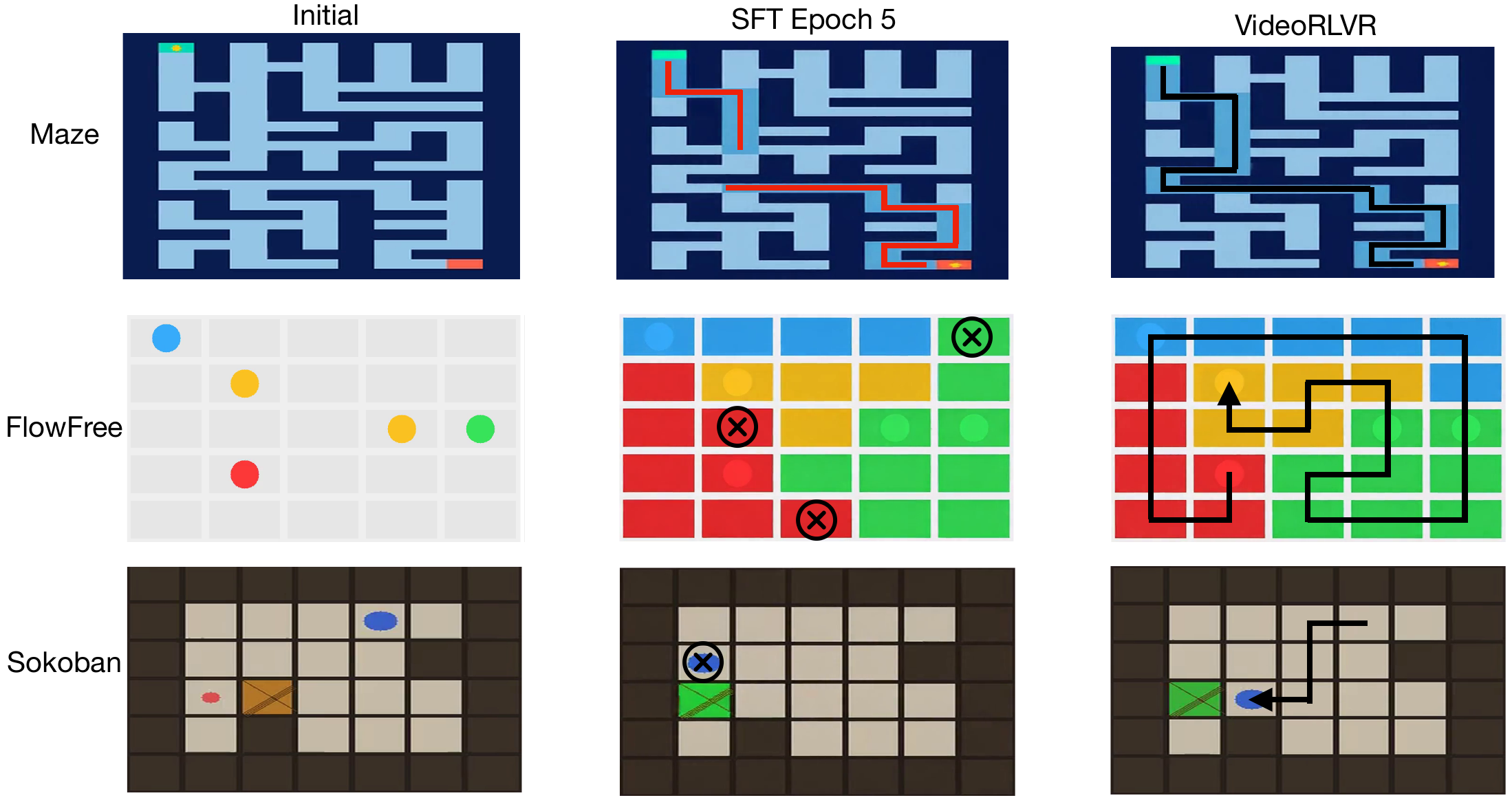}
    \vspace{-6mm}
    \caption{Qualitative case study across three reasoning domains. We compare generations from the SFT baseline (Epoch 5) and the VideoRLVR model.}
    \label{fig:case_study}
    \vspace{-5mm}
\end{figure}

\subsection{Case Study}
\Cref{fig:case_study} provides qualitative examples across Maze, FlowFree, and Sokoban.
The SFT baseline often captures the visual format of each domain, such as drawing paths, coloring grids, or rendering objects, but it can fail to satisfy the task rules.
For example, SFT outputs may contain disconnected paths, inconsistent color connectivity, or invalid object transitions.
In the Sokoban example, the SFT model produces a visually plausible but invalid shortcut rather than a valid sequence of box-pushing actions.
In contrast, the VideoRLVR-optimized model more consistently satisfies the symbolic constraints checked by our verifiers.
Across the shown examples, it produces connected paths, more coherent grid solutions, and more valid object transitions while preserving the overall visual structure of the task.
These qualitative results support the quantitative findings: verifiable RL improves rule-based correctness beyond what is achieved by supervised imitation alone.
\section{Conclusion}
\label{sec:conclusion}
This work studies reinforcement learning with verifiable rewards for video reasoning and introduces \textbf{VideoRLVR}, a practical recipe for optimizing video reasoning models. 
By combining rule-verifiable data generation, an SDE-GRPO optimization backbone, dense decomposed rewards, and Early-Step Focus, VideoRLVR addresses the gap between perceptual video synthesis and task-level logical correctness. 
Our experiments show that supervised fine-tuning provides an important visual and structural prior, but can plateau or degrade on harder reasoning tasks when optimized only through imitation. 
In contrast, verifiable RL improves success rates across Maze, FlowFree, and Sokoban, with dense decomposed rewards proving especially useful in low-success-rate domains. 
We further show that Early-Step Focus reduces training time by about 40\% with little observed loss in reasoning performance. 
Overall, our results suggest that verifiable RL can substantially improve the logical correctness of generated videos, enabling open-source video models to outperform stronger general-purpose video generation baselines on visual reasoning benchmarks.


\newpage
{
\bibliographystyle{plain}
\bibliography{reference}
}

\newpage
\appendix
\section{Dataset Generation Details}
\label{sec:appd_dataset}

We generate all training and evaluation instances using rule-based algorithms so that each sample has a known valid trajectory and task metadata for automatic verification. 
Each generated instance consists of an initial visual state, a textual task instruction, a ground-truth state/action trajectory, and a rendered video. 
The generation process differs by task, but all domains are designed so that retained samples have verified valid solutions.

\subsection{Maze Generation}

For Maze, we first generate a connected maze graph using standard maze-carving algorithms, including depth-first search, Prim's algorithm, and Kruskal's algorithm. 
Given a generated maze, we set the logical start and goal cells to opposite corners, $(0,0)$ and $(n-1,n-1)$, respectively. 
The ground-truth trajectory is obtained by running shortest-path search on the maze graph from the start to the goal. 
This produces a sequence of adjacent logical cells that forms a valid path without crossing walls.

To render this logical trajectory as a continuous visual path, we expand logical cell coordinates into pixel-level path coordinates. 
For every move between adjacent cells, we insert the corresponding corridor cell between them, producing a continuous walkable sequence in the rendered maze. 
Step directions are then derived by differencing successive coordinates, yielding a sequence of discrete actions in $\{\mathrm{U},\mathrm{D},\mathrm{L},\mathrm{R}\}$. 
This action sequence is used to render the ground-truth video and to support verifier-based evaluation of connectivity and wall consistency.

\subsection{FlowFree Generation}

For FlowFree, we generate the solution first and derive the puzzle from it. 
We construct a Hamiltonian path over the $n\times n$ grid using Warnsdorff-style search: at each step, the algorithm prioritizes neighboring cells with the fewest onward unvisited moves, with random tie-breaking and bounded retries. 
This produces a full-coverage path over the grid.

We then partition the Hamiltonian path into $k$ contiguous segments, where $k$ is the number of colors. 
The split indices are sampled randomly subject to the constraint that each segment contains at least two cells. 
For each segment, the two endpoints become the colored endpoint dots shown in the puzzle, while the full segment is retained as the ground-truth flow for that color. 
Because the puzzle is constructed from the solution trajectory, every generated instance has a known valid solution by construction. 
The verifier can therefore check endpoint preservation, color connectivity, non-overlap, and grid coverage against the stored segment metadata.

\subsection{Sokoban Generation}

For Sokoban, we first synthesize a candidate level by sampling connected floor cells, player position, box positions, and target positions. 
During sampling, we apply basic deadlock filtering to avoid trivially unsolvable configurations, such as boxes placed in wall corners where they cannot be pushed out. 
Candidate levels are then passed to a symbolic Sokoban solver.

The solver performs breadth-first search over the discrete state space. 
Each state is represented by the player position and the set of box positions, written as $(p, B)$ where $p$ is the player cell and $B$ is a set of occupied box cells. 
At each expansion, the solver considers the four possible player moves. 
If the target cell is empty floor, the player moves without changing the box set. 
If the target cell contains a box, the move is valid only when the cell beyond the box is also empty floor; in that case, the box is pushed one cell forward. 
The goal condition is satisfied when the set of box positions matches the set of target positions.

Because breadth-first search explores states in increasing action length, the first returned solution is an optimal-move trajectory under this transition model. 
To control generation cost, we cap the number of expanded states. 
Candidate levels that cannot be solved within the cap are discarded and regenerated. 
Thus, every retained Sokoban sample has a verified valid trajectory, along with process-level metadata for checking player motion, box motion, illegal pulls, teleportation, and final target satisfaction.



\section{Experimental Setup}
\label{sec:appd_experimental_setup}

\subsection{Dataset}
\label{sec:appd_dataset_setup}
\vspace{-2mm}
\begin{itemize}[leftmargin=1.5em,itemsep=0pt]
    \item \textbf{Maze:} We generate $10,000$ samples with grid dimensions ranging from $7{\times}7$ to $21{\times}21$. Each instance pairs an unsolved layout containing start and goal markers with a ground-truth video that renders the contiguous path between them.
    \item \textbf{FlowFree:} We generate $10,000$ puzzles with grid sizes between $5{\times}5$ and $8{\times}8$ by splitting Hamiltonian paths into colored segments, ensuring that a valid solution must occupy all available cells. The initial frame displays only the discrete color-pair endpoints, and the video unrolls the progressive coloring of each path.
    \item \textbf{Sokoban:} This domain includes $10,000$ puzzles with grid sizes from $6{\times}6$ to $10{\times}10$ and $1$--$3$ boxes. Solution trajectories are capped at $60$ moves. The input frame depicts the initial board configuration, while the video unrolls the solution at a resolution of one agent push per frame, ensuring strict alignment between temporal and logical steps.
\end{itemize}
\vspace{-2mm}

\subsection{Evaluation}
\label{sec:appd_eval}
\stitle{Trajectory Alignment Metrics}
To measure the alignment with the ground-truth reference, we compute precision, recall, and F1 at the unit most natural to each task's solution manifold:
\begin{itemize}[leftmargin=1.5em,itemsep=0pt]
    \item \textbf{Maze (Pixel-level):} We compute a change mask between the initial and final frames to isolate the generated path from the static background. This ensures the metric captures the model's intervention rather than background reconstruction.
    \item \textbf{FlowFree (Cell-level):} We extract mean colors for each grid cell in the terminal frame. This avoids penalizing anti-aliasing artifacts and focuses on the semantic correctness of the path coloring.
    \item \textbf{Sokoban (Action-level):} Since multiple trajectories can yield the same final state, we decode the video into a symbolic action sequence $a \in \{\mathrm{U\text{(Up)},D\text{(Down)},L\text{(Left)},R\text{(Right)}}\}$. We report position-aligned F1, which penalizes out-of-order or invalid moves.
\end{itemize}

\stitle{Symbolic Success Rate}
Alignment with a single GT reference is insufficient to detect valid alternative solutions or visually plausible but rule-violating outputs. We therefore implement binary success detectors that parse the video $\mathbf{V}$ into symbolic states to verify task-specific rules:
\begin{itemize}[leftmargin=1.5em,itemsep=0pt]
    \item \textbf{Maze Success:} Requires a connected path between markers without violating wall constraints.
    \item \textbf{FlowFree Success:} Validates endpoint preservation, color connectivity, and the grid fill-rate.
    \item \textbf{Sokoban Success:} Evaluates \textit{process validity} over all frames, checking that player and box displacements follow physics-based rules and the final state matches the target.
\end{itemize}

\begin{figure}[t]
    \centering
    \includegraphics[width=\linewidth]{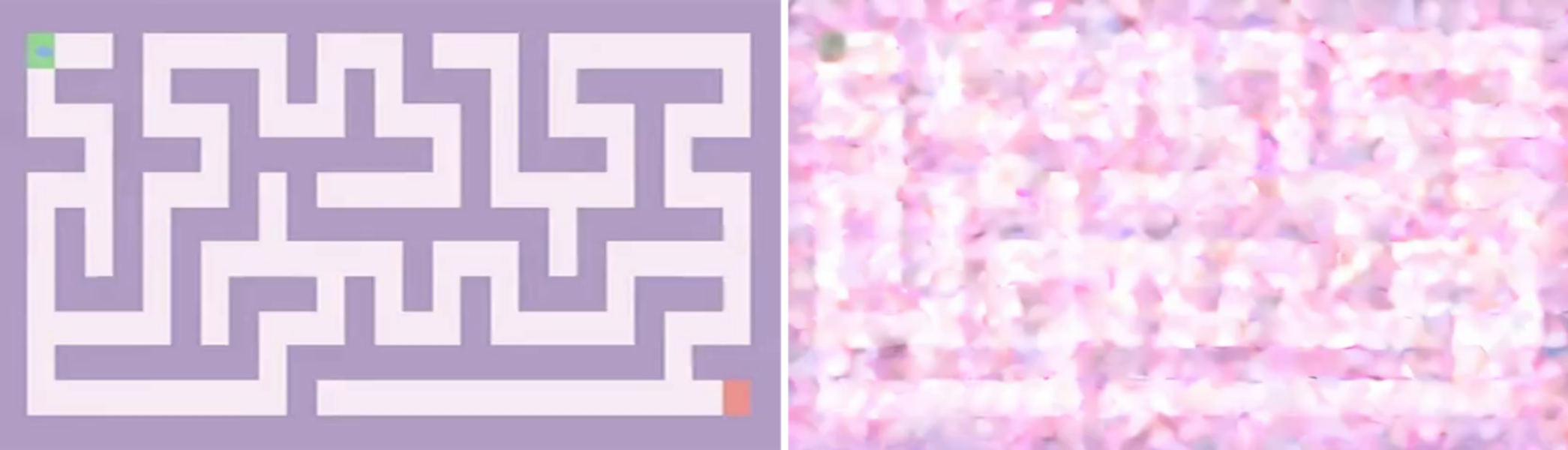}
    \caption{Qualitative example of reward hacking in the absence of KL-divergence regularization. It preserves the wall constraint, and saturates all paths to connect two endpoints, thereby achieving a maximal reward.}
    \label{fig:KL_ablation}
\end{figure}

\section{Analysis}

\subsection{KL Constraint}
\label{sec:appd_kl}
To study the role of KL regularization, we compare VideoRLVR training with and without the KL penalty while keeping all other hyperparameters fixed. 
Both runs start from the same SFT checkpoint, use the same group size, denoising-step budget, reward function, and training prompts, but differ in the KL coefficient $\beta$. 
The regularized setting uses the default value $\beta=0.04$, while the ablated setting sets $\beta=0$, removing the constraint to the SFT reference policy. 
We inspect generations from an early stage of GRPO training, where policy drift and reward-hacking behavior are most visible. 
This controlled comparison isolates the effect of the KL term on preserving the model's visual prior during verifier-driven optimization.


\end{document}